# An Intraoperative Force Perception and Signal Decoupling Method on Capsulorhexis Forceps

Heng Zhang, *Student Member, IEEE*

*Abstract*—**Force perception on medical instruments is critical for understanding the mechanism between surgical tools and tissues for feeding back quantized force information, which is essential for guidance and supervision in robotic autonomous surgery. Especially for continuous curvilinear capsulorhexis (CCC), it always lacks a force measuring method, providing a sensitive, accurate, and multi-dimensional measurement to track the intraoperative force. Furthermore, the decoupling matrix obtained from the calibration can decorrelate signals with acceptable accuracy, however, this calculating method is not a strong way for thoroughly decoupling under some sensitive measuring situations such as the CCC. In this paper, a three-dimensional force perception method on capsulorhexis forceps by installing Fiber Bragg Grating sensors (FBGs) on prongs and a signal decoupling method combined with FASTICA is first proposed to solve these problems. According to experimental results, the measuring range is up to 1 N (depending on the range of wavelength shifts of sensors) and the resolution on x, y, and z axial force is 0.5, 0.5, and 2 mN separately. To minimize the coupling effects among sensors on measuring multi-axial forces, by unitizing the particular parameter and scaling the corresponding vector in the mixing matrix and recovered signals from FastICA, the signals from sensors can be decorrelated and recovered with the errors on axial forces decreasing up to 50% least. The calibration and calculation can also be simplified with half the parameters involved in the calculation. Experiments on thin sheets and in vitro porcine eyes were performed, and it was found that the tearing forces were stable and the time sequence of tearing forceps was stationary or first-order difference stationary during roughly circular crack propagating.**

## I. INTRODUCTION

THE human eye has complex and precise optical working principles. Cataracts are serious eye diseases and one of the main factors that can lead to a fierce decline in human eyesight or vision loss[1]. The main pathogenic factor of cataracts is protein denaturation of the lens. Therefore, the transmission of light through the lens gets worse and the vision of patients becomes blurred. The curing method is replacing the diseased lens with an artificial lens at an appropriate position. The position should guarantee the focus of an artificial lens stays close to the eye's focus and is mainly decided by appropriate manipulations performed by surgeons in the continuous circular capsulorhexis (CCC). The CCC is one essential procedure to realize intraocular lens replacement in

Heng Zhang are with the School of Electrical and Information Engineering, Zhengzhou University, Zhengzhou 450001, China. (e-mail: zh2501365172@163.com; zh2501365172@gs.zzu.edu.cn).

cataract surgery, requiring precise manipulations under the microscope[2]. This procedure is about using forceps to form a smooth, circular crack in the anterior capsule and remove the detached part of the capsule, so a circular wound for exchanging the diseased lens would appear. If the crack is not circular enough, the artificial lens will not be in the right position to fulfill its optics function. The continuous circular capsulorhexis (CCC) is developed in medical practices with no sufficient theoretical description of its mechanisms[3]. The results of the CCC heavily rely on the expert experience and skills of surgeons, which set limits to promote and popularize the CCC for a large amount of time for practicing and training. There must be some regularities in intraoperative interaction forces, postures, and movement of the surgeon's hand and tips of forceps, which can lead to a well-circular tearing wound on the anterior capsule. It can be assumed the factors that influence the propagation of cracks are the trajectory of the instrument and the interaction forces between forceps and the anterior capsule of the eye. Getting to know these regularities described in physics, dynamics, and numerical information is likely to help a lot in quantizing expert experience, training, surgical standardization, and robotic surgery. Therefore, it is necessary to develop a force sensing method on forceps to gather force information in CCC operation.

Force perception on forceps can collect intraoperative interaction force information, which can help greatly in analyzing and understanding what happened during operations rather than surgeons' ambiguous skill feelings. In CCC, the real-time fluctuation of force is likely to be related to the state of the capsule's crack propagation. The interaction force information between the tips of forceps and the tissue, including directions and amplitude on XYZ axes, will inspire the exploration of what factors determine crack propagation of the anterior capsule. However, there is still not much choice of high-resolution, three-axial methods to measure forces between tissues and tips of capsulorhexis forceps. Until now, some force-sensing methods on instruments have been discussed in previous studies. In the work of Zareinia et al[4], pairs of strain gauges were implemented on each side of prongs of bipolar forceps, and the planar forces between the tips of forceps and brain tissues can be recorded in real-time. The strain gauges can be 1.6 mm×1.2 mm, which is suitable for slim prongs. This work indicated a practical method to obtain force information by building the relationship between the strains and forces. Chen et al[5] proposed an ultrathin pressure feedback sensor for intraventricular neurosurgery robotic tools. This sensor is soft. Surgical tools can be wrapped by the sensor so that the collision with the brain tissues can be detected. The soft piezoresistive sensor had also been applied on retractors, but it was not stable enough and could not detect static forces[7]. By the way, the planar force information or pressure information is not adequate. Centered on multi-dimensional force information, Marcus et al[8] estimated the forces exerted on the surgical instrument during micro neurosurgery by a Quanser 6 Degrees-Of-Freedom Telepresence System integrated into the robot arm. The haptic device in the system has the capability of providing 6 DOF force/torque feedback to the operator. This method is about proving force feedback from the end of the tool like a probe with a resolution of 0.01N. This work is mainly about estimating forces exerted during cranial microsurgery, the force sensing method is very direct. By using 6 DOF force/torque sensors, full dimensional force information can be obtained. For more detailed information, strain gauges and six-component force sensors like nano-17 were first applied on bipolar forceps in neurosurgical scenarios by the works of Maddahi et al[9]. The six-component force sensor is not a micro-sensor so was integrated with surgical tools as the end effector of the Neuroarm[10]. Therefore, these instruments can be complex devices rather than handed medical tools. To meet the request for small-size sensors, Kang et al[11] developed a bioresorbable silicon electronic sensor with a size of 1mm×2mm×0.08mm to detect pressure in intracranial pressure. Shin et al[12] also developed optical sensor realization. However, these sensors are highly specialized and the sensors are expensive, and cannot be easily manufactured. The small size also cannot have a large measuring capacity. Considering the sensitivity, capacity, size, and biotoxicity of sensors, Zhang et al.[13][14] first estimated force-measuring performance on bipolar forceps by installing pairs of FBGs.

Generally speaking, there is not a general sensing method, which can suit all the instruments and measuring occasions. It is about the consideration of sensors (safety, size), instruments (structure, size), and surgical scenarios. Force perception on forceps is one of the concerns in this paper, a proper force sensing method can collect intraoperative interaction force information, which can help greatly in analyzing problems. Forceps is a classical structure of medicine tools, and its structure is very simple and useful. In CCC, sensors must be small enough to be adapted to quite limited space of the anterior chamber so that would not disturb forceps' works. The structure of forceps is typical, simple, and practical. Therefore, it is not likely for a lot of modifications to the structure of forceps. Thus, the force perception method would better meet the next requirements: 1. The Least modifications on the structure of forceps. The least influence on the working function of forceps. 2. Small in size and can be adjusted to the commonly used surgical instrument. 3. The resolution of axial force is better than 0.01 N. The measuring signals can be obtained by a designed force perception device. It is also essential to recover signals from observed signals from the force perception device. Therefore, another important part of this study is decoupling sensor signals in a way with good accuracy, so the decoupled data can be well estimated and the real tendency can be reflected. However, signal processing is not discussed in detail in above mentioned studies. The common method mentioned in previous works[4][13] realized signal decoupling mainly by calibration. However, the calibration method mentioned above is not a good enough decorrelation method, especially for the occasion with a resolution reaching milli-newton in this study. In application, the sensors on instruments measure the coupling strains caused by the resultant force. Because the sensors are sensitive to different axial forces. In the calibration process, the linear relationship between force and signals is decoupled by a coefficient matrix. The matrix can describe the influence of axial forces on sensors. The coefficients in the matrix can describe force response on sensors by algebra calculation rather than separate different components in signals. For the calibration operations and matrix calculation, errors exist and cannot be erased, even finally amplified, resulting in bad force results. These errors accumulated in calculation and calibration may not be obvious in some structure on a large scale but plays a main role in force measuring with some very sensitive structure, such as very slim beam, like prongs of capsulorhexis forceps with minimum width below 1 mm. Sensitive force perception on slim beams may be changed for the variation of boundary conditions and calibration operation. Therefore, the calibration matrix obtained from calibration may not be reliable for more sensitive measurement occasions. The problem is that the matrix obtained from calibration cannot recognize the components of observed signals from sensors. The x, y, and z axial forces are independent, and the corresponding independent component belongs to each axial force among the observed signals. There are some regularities of the problem: 1. The relationship between axial force and signals follows a linear relationship. 2. The axial forces on the XYZ axes are independent of each other. 3. The different sensor is sensitive to the different axial component of force. 4. The number of independent components can be equal to the number of sensors.

For separating components of signals, it is natural to use independent component analysis (ICA) to separate these components. The FastICA was proposed in Hyvälinen 's work[15]. As a strong and fast data recovering method to separate linearly mixed independent components from the observed signals getting from the sensors. However, although the true fluctuations of components can be obtained from ICA, there are still some inherent ambiguities that need to be solved. The ambiguities are [16]: 1. Sign ambiguity problem of independent components. 2. The variances (energies) of the independent components cannot be determined. 3. The order of independent components cannot be determined. Some studies once had focused on solving these ambiguities so that the true values of components of signals could be recovered. In the work of Sawada et al[17], the permutation problem of multiple source signals is solved based on the direction of arrival estimation and the inter-frequency correlation of signal envelopes. Alvarez et al[18] also compared the performance of SOBI, FastICA, JADE, and Informax algorithms. The FastICA is also applied to real data from a gravitational wave detector for the first time. With more restrictions from reality, the real data of gravitational waves can be well recovered[19]. Inspired by the above works, based on the FastICA, the ambiguities of unmixed signals can be corrected with more restrictions from the physical characteristics of signals. Therefore, with some skills in data preprocessing and modifications on FastICA, better decoupling results of decoupling signals from two pairs of sensors can be obtained. As for experiments, better-calculated results of forces revealed some interesting phenomena.

Motivated by the above considerations, the features and contributions of the works are listed as follows.

1. firstly, to our best knowledge, a three-dimensional force perception method on capsulorhexis forceps was first established to solve a lack of muti-axis force information collective device in CCC. A novel FBG layout and force-resolving strategy with high sensitivity is proposed. This strategy is about taking advantage of the bending of the prong, applying pairs of FBGs on perpendicular surfaces of different positions near the bending part of the prong to detect the planar forces in different coordinates. Because the difference of y axial forces in the two coordinates indicates the component of the z axial force. Therefore, the z-axial force can be measured and all three-dimensional forces can be tracked benefiting from the strategy.

2. Secondly, the phenomenon was observed that the coupling inside signals can be worse for sensitive force measurements by multiple sensors, which always leads to poor performance on decoupling signals because of significant residuals. To minimize these errors, a method combined with FastICA for signal processing is proposed for decorrelating light wavelength shift signals of sensors and recovering the real force exerted on tips of forceps. The method was mainly about unitizing the particular parameter, scaling the corresponding vector in the mixing matrix, and synchronously compensating signals from FastICA, and the signals from sensors can be well decorrelated and recovered with the errors on axial force significantly decreasing. Meanwhile, the method can significantly simplify the calibration and the calculation process because only half the coefficients are necessary compared with the previous method.

3. Thirdly, it was found that the tearing forces that occurred during the propagation of the nearly circular crack were stable, and the time sequence of tearing forceps was stationary or first-order difference stationary in some operations in experiments, which indicates that the forces are likely to be predicted and trackable. These features can inspire further research on the relationship between the circular crack, the trajectory of forceps, and interaction forces.

The remainder of this paper is arranged as follows. In section II, the operation of CCC, the manipulations of forceps, and the interaction of tips of forceps with exerted forces are analyzed. In section III, the scheme of the force perception method on forceps is illustrated, including simulation, and mechanics modeling on forceps. In section IV, the calibration results and the experiment results are presented. In section V, it is explained in detail how the signals obtained from sensors are decorrelated and the force information was recovered. Finally, the conclusion is given in Section VI.

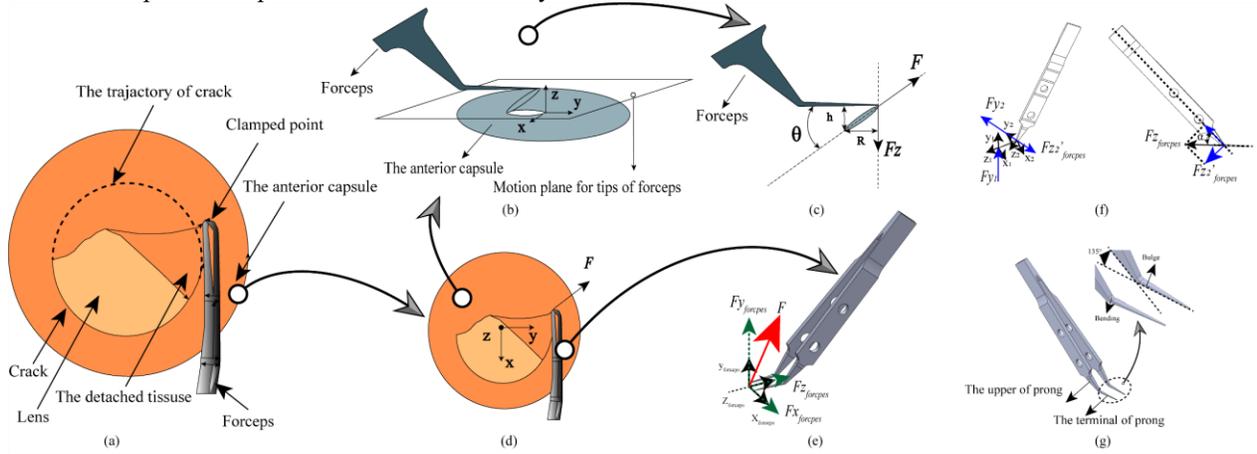

**Fig. 1.** Scheme of the CCC and working principles of forceps. (a) The illustration for operations of the CCC, (b) The movement of the forceps, (c) $F_z$ loaded on tips of forceps, (d)Illustration of force in CCC, (e) The force decomposition in the coordinate set on tips of forceps, (f) Illustrations of forces decomposition in coordinates set on a prong of forceps, (g) The geometric configuration of capsulorhexis forceps.

## II. FORCEPS OPERATION ANALYZING

### A. Operation analysis of CCC and forceps

As shown in **Fig. 1**(a), the key to a successful CCC is to form a good circular crack on the anterior capsule. At the clamped point, one piece of the anterior capsule is stretched with the movement of the tips of forceps. Until satisfying some criteria (can be the occurrence of Crack-tip Stress Field, reaching the maximum strains, according to fracture mechanics on thin sheets) for the propagation of crack on the anterior capsule, the wound will appear. It can be assumed that the propagation of the crack can be attributed to the trajectories of the clamped point and the stretching force on tissue from the tips of forceps. In this paper, our scope is mainly exploring the force information during the CCC. The movement of forceps is described by (b) in **Fig. 1**. For the limited space of the anterior chamber of the eye, the movement of tips can be seen as restricted in a plane. When using forceps to drag the capsule, a pulling-like force occurs on the tips of the forceps. The coordinate setup is as shown in **Fig. 1** (b), (c), (d), and (e). The coordinate $X_{froceps} Y_{froceps} Z_{froceps}$ named B is set for describing force calculating from sensors on forceps. The coordinate XYZ is named A. The forces on tips can be described as (1):

$$F = {}^B_A R \, F_{forceps} = {}^B_A R \begin{bmatrix} F_{X_{forceps}} \\ F_{Y_{forceps}} \\ F_{Z_{forceps}} \end{bmatrix} \qquad (1)$$

In which $F$ is the exerted force, ${}^B_A R$ is the transformation matrix, and $F_{forceps}$ is the exerted force in the coordination of forceps. From the (b) and (c) in **Fig. 1**, one of the components of force expressed in the coordinate of the eyes is $F_z = F sin\theta$, where the $F_z$ is the pulling-down force on the tips of forceps described in coordinate XYZ, $F$ is the resultant force between forceps and tissues, $\theta$ is the angle between the prong and detached part of the anterior capsule. It can be seen that $\theta$ is small for the ratio of $\frac{h}{R}$ is near zero ($\frac{h}{R} \approx 0$). Thus, the $sin\theta \approx$ 0 can be established for approximation in physics. Therefore, $F_z$ takes up much less percentage than other components of force. The structure of forceps is important for the force perception of forceps shown in **Fig. 1**(g). The forceps have a unique structure, which makes them different from the other types of surgical forceps. As shown in **Fig. 1**(g), the prongs of forceps are bent near the tips of forceps rather than straight. The bending angle is approximately 135°. And there is a bulge at the inner side of the bending part. The prong is slim with a minimum width below 1 mm. Therefore, the prong is sensitive to forces and also easily fails under excessive loads. This bulge can keep the stress applied on prongs in a safety range, which is necessary to prevent the prongs from deformation under excessive loads from the operator's hand.

Table I

SUMMARY OF FORCE PERCEPTION ON MEDICAL INSTRUMENTS

| Group | Range | Resolution | | | Dimensions | Instrument | Application Scenario |
|---|---|---|---|---|---|---|---|
| Zareinia et al[4] | [-1.5,1.5] N | 10 mN on each axis | | | 2 | Bipolar forceps | Neurosurgery |
| Zhang et al[14] | [0,4] N | 0.1 N on the x-axis | | 0.03 N on the y-axis | 2 | Bipolar forceps | Neurosurgery |
| Zhang et al[13] | [0,4] N | 0.01 N on x-axis | 0.03 N on the y-axis | 0.1 N on the z-axis | 3 | Bipolar forceps | Neurosurgery |
| Iordachita et al[20] | — | 0.25 mN（planar force） | | | 2 | Microsurgical hook | Retinal microsurgery |
| Zheng et al[21] | — | 0.83mN（planar force） | | | 2 | Microsurgical forceps | Cataract surgery |
| Our work | > 1 N | 0.5 mN on the x-axis | 0.5 mN on the-axis | 2 mN on the z-axis | 3 | Capsulorhexis forceps | CCC in cataract surgery |

## III. FORCE PERCEPTION METHOD ON FORCEPS

### A. Forceps perception method on forceps

The three-dimensional force perception scheme on the capsulorhexis forceps is shown in **Fig. 2**.

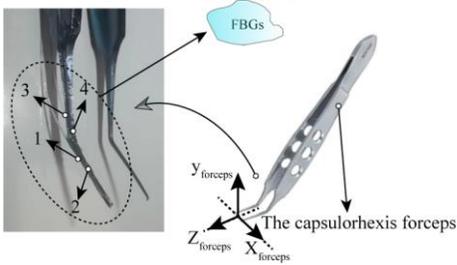

**Fig. 2.** The scheme of force perception on forceps.

FBGs are employed in this study to suit the slim prong of forceps mainly for the small size and sensitivity to strains. The arrangement of FBGs on prongs is shown in **Fig. 2**. Pairs of FBGs were installed in different positions separately on the surface of each prong. In this study, sensors on one prong are enough. The principle of force-sensing is measuring strains on prongs by sensors. The light wavelength shifts from sensors are the observed signal and measurement of strains, and they follow a linear relationship. From the numerical relationship between force and light wavelength shifts, the exerted force on the tips of forceps will be recovered. In Table I, a summary of related works and a contrast of parameters are given to illustrate the specialty of the forceps perception method proposed in this paper.

### B. Mechanics modeling on capsulorhexis forceps

As proved in previous works[4], the prong of forceps can be modeled as a cantilever beam. Thus, the deformation along the prong can be expressed in (2):

$$w_b = -\frac{Fl^3}{3EI} \qquad (2)$$

Where $w_b$ is the deflection of the prong, $F$ stands for the, $l$ is the length of the beam, $E$ is Young's modulus, and $i$ is the inertia of the cross-section along the prong. For axial force loading occasions, the relationship between wavelength shifts on FBGs and axial force can be expressed:

$$F = k\Delta w + b \qquad (3)$$

Thus, the deformation of the prong can be calculated by the light wavelength shift $\Delta w$ of sensors:

$$w_b = -\frac{(k\Delta w + b)l^3}{3EI} \qquad (4)$$

According to the mechanics of elastic materials, the deformation of elements occurred not in a single direction because of the poison's effect. It implies that strains are coupled. The relationship between stress and strains for elastic

elements are:

$$\begin{bmatrix} \sigma_x \\ \sigma_y \\ \sigma_z \end{bmatrix} = \frac{E}{(1+v)(1-2v)} \begin{bmatrix} (1-v) & v & v \\ v & (1-v) & v \\ v & v & (1-v) \end{bmatrix} \begin{bmatrix} \varepsilon_x \\ \varepsilon_y \\ \varepsilon_z \end{bmatrix} \quad (5)$$

The relationship between light wavelength shifts can be built as mentioned in Zhang's previous work[13]:

$$\begin{bmatrix} w_1 + \Delta w_1 \\ w_2 + \Delta w_2 \\ w_3 + \Delta w_3 \end{bmatrix} = \begin{bmatrix} k_{11} & k_{21} & k_{r31} \\ k_{12} & k_{22} & k_{r32} \\ k_{13} & k_{23} & k_{r33} \end{bmatrix} \begin{bmatrix} F_{xr} \\ F_{yr} \\ F_{zr} \end{bmatrix} + \begin{bmatrix} w_{ro1} \\ w_{ro2} \\ w_{r03} \end{bmatrix} \quad (6)$$

However, it cannot be directly applied to the forceps discussed in this paper: 1. The terminal of the prong is very slim, and cannot support installing too many sensors. 2. It will be strict for operations in the calibration process, especially for components of force along the z-axis.

The occasion on which there is only exerted force along the z-axis is special and needs to be discussed. On this occasion, if there is no bending of the terminal part of the prong, only the compression exists inside elements on the prong. The strains that happened in pure compression can be described as:

$$\varepsilon = \frac{F_{z\,forceps}}{EA} \quad (7)$$

Where $\varepsilon$ is the strain, $F_{z\,forceps}$ stands for the compressing or pulling force along the $z_1$. And reasons for suggesting not using (8) and (9): 1. Most of the metals that medical instruments are composed of, are difficult to compress. Thus, the resolution of measuring the z-axial component of force would not be high. 2. It brings great difficulties in the calibration of the axial force along the z. Because the prong can be easily bent, the relationship between the compression force

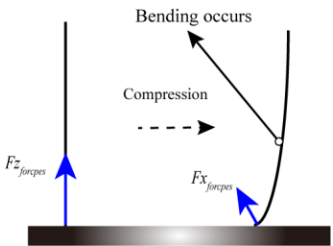

**Fig. 3.** Scheme of prong's bending under z-axial component force: the calibrating $F_{z\,forceps}$ in fact becomes calibrating $F_{x\,forceps}$.

The special structure can be useful in realizing the detection of the X, Y, and Z axial components of force. The planar force perception by using strain gauges and FBGs has been proven reasonable [12]. Therefore, the planar force sensing can be achieved by each pair of FBGs, and the principle of estimating $F_{z\,forceps}$ is shown in **Fig. 1**(f). The sensors at different parts of the prong monitor the planar forces at coordinate x1y1 and coordinate x2y2 separately. For the bending of the prong, if the planar force measuring is accurate, then, $F_{x_1} = F_{x_2}$. When $F_{z\,forceps}$ is applied, there are only responses corresponding to $F_{z2\,forceps}$ from sensors 3 and 4.

$$F_{z\,forcpes} \sin\alpha = F_{z2\,forcpes} = F_{y2} - k_n F_{y1} \cos\alpha \quad (8)$$

Therefore, the difference from $F_{y2}$ and $F_{y1}$ can tell the $F_{z\,forceps}$. Forces perception on forceps is mainly about measuring linearly responded strains on beams caused by applied forces. The main idea of measuring loads on tips of forceps is about building a relationship between forces and sensor signals. For the linear addition characteristic of strains for elastic elements, the relationship between strains and stress is:

$$\begin{bmatrix} \sigma_x \\ \sigma_y \end{bmatrix} = \frac{E}{(1+v)(1-2v)} \begin{bmatrix} (1-v) & v \\ v & (1-v) \end{bmatrix} \begin{bmatrix} \varepsilon_x \\ \varepsilon_y \end{bmatrix} \quad (9)$$

Where $\sigma_i$ denotes the stress along the $i$-axis, $\varepsilon_i$ stands for the strain on the $i$-axis, and $v$ is the Poisson ratio. Modifications with light wavelength shifts from sensors:

$$\begin{cases} \begin{bmatrix} w_1 + \Delta w_1 \\ w_2 + \Delta w_2 \end{bmatrix} = \begin{bmatrix} k_{11} & k_{21} \\ k_{12} & k_{22} \end{bmatrix} \begin{bmatrix} F_{x1} \\ F_{y1} \end{bmatrix} + \begin{bmatrix} w_{o1} \\ w_{o2} \end{bmatrix} \\ \begin{bmatrix} w_3 + \Delta w_1 \\ w_4 + \Delta w_2 \end{bmatrix} = \begin{bmatrix} c_{11} & c_{21} \\ c_{12} & c_{22} \end{bmatrix} \begin{bmatrix} F_{x3} \\ F_{y3} \end{bmatrix} + \begin{bmatrix} w_{o3} \\ w_{o3} \end{bmatrix} \\ F_{z\,forceps} = \frac{F_{y2} - k_n F_{y1} \cos\alpha}{\sin\alpha} \end{cases} \quad (10)$$

Thus, the relationship between forces and light wavelength shifts $\Delta w_i$ is linear. The accuracy of $F_{yi}$ can have an impact on the accuracy of $F_{z\,forceps}$.

## C. Simulation of mechanics on forceps

FEM simulation of the mechanics of forceps is necessary for figuring out the distribution of strains along the prong and where the optimal strains occur, which is helpful to decide the position for installing sensors for getting signals with high Signal to Noise Ratio (SNR), and estimating the performance of the force perception scheme. Because the prong of the forceps is sensitive to the exerted forces, the change of boundary conditions might cause unstable results in calibration. Thus, another reason for simulation is to estimate the influence of changes in fixed points on calibration. Therefore, a 3D digital model of forceps was sketched and stimulation sections were activated to estimate the distribution of strains along forceps and analyze the influence of fixed position on strains (see **Fig. 4**) in Solidworks. The material was Titanium alloy. The Young's modulus ($E$) is $1.048 \times 10^3$ Mpa. The density is 4428.784 kg/m³ and the poison ratio is 0.31.

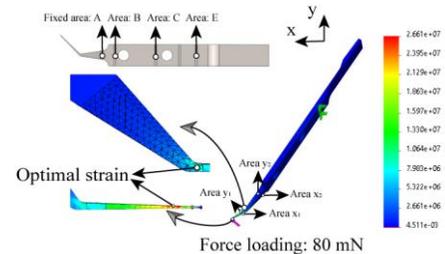

Force loading: 80 mN

**Fig. 4.** Strain contour plot along one prong.

As shown in **Fig. 4**, optimal strains occurred in front of the prong. The strains decline along the prong from the tips generally. Therefore, sensors should be installed as close to bending and the tips of forceps. The resolution of each axial component of the sensors is in Table I. It can be supported by the FEM simulation that the relationship between strains and exerted forces follows a linear pattern on every point along the prong with varying cross-sections of the prong. Therefore, the total strains along the prong in an area can be described as:

$$\varepsilon_{areai} = \oint f_p \, strains \tag{11}$$

Where the $\varepsilon_{areai}$ are the total strains in a certain area on the prong. For sensors, it stands for the detecting area. The $f_p \, strains$ is the function, which stands for the strain's distributions along the prong. The influence of changes on the fixed area is shown in **Fig. 5**, it can be seen that the changes in the boundary condition of the fixed area would not significantly cause great changes in strains along the prong.

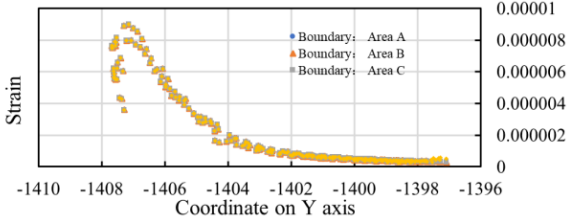

**Fig. 5.** The distributions of strains stay the same when the fixed area changes.

Table II
RESOLUTION OF AXIAL FORCE ON FORCEPS

| The axial components of Force | Sensitivity from measuring/mN | Sensitivity of force from simulation/mN |
|---|---|---|
| X | 0.5 | ≈0.5 |
| Y | 0.5 | ≈0.5 |
| Z | 2 | 5 |

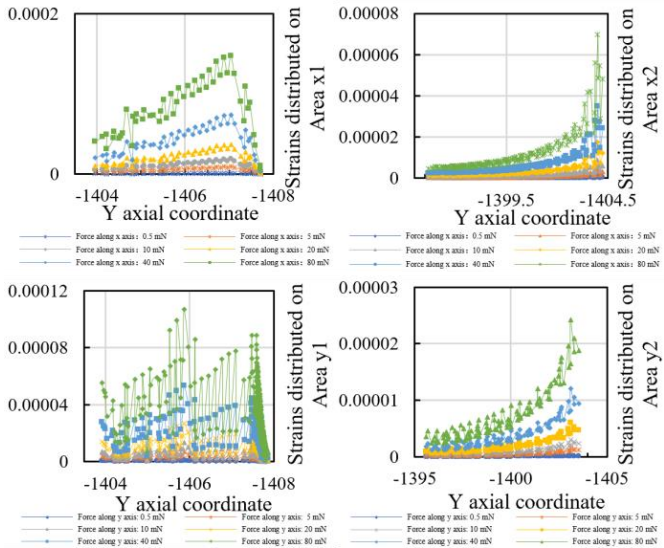

**Fig. 6.** Distribution of strains along one prong.

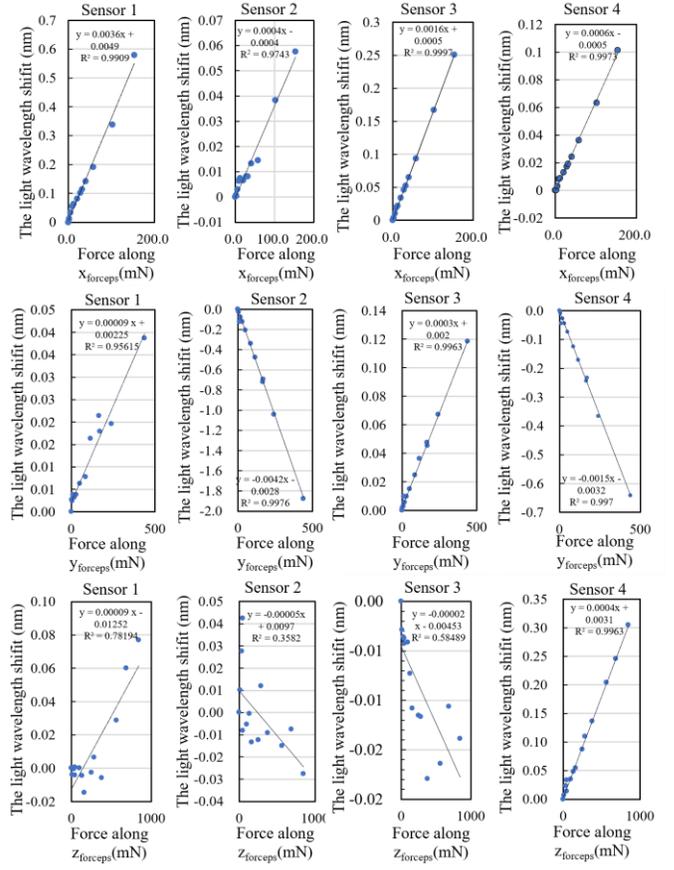

**Fig. 7.** Calibration results of sensors for axial forces.

### D. Modifications on forceps

The bolt and nut had been added to the forceps, and the ring in the middle works as a new bulge to prevent excessive forces.

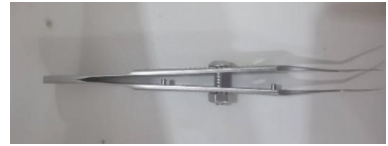

**Fig. 8.** Modifications on forceps. A limit part was added in case excessive forces were applied.

## IV. SIGNALS ANALYZING AND PROCESSING

### A. The shorting of the calibration method

The relationship between force and signals of sensors can rely on the calibration calibrations mentioned above. The more accuracy the situations require, the more complicated and restricted the operations are, especially for calibrating $F_{z_{forceps}}$. It is also not a strong way to decorrelate signals and is likely to lead to distinct errors in the true values of axial forces. As shown in **Fig. 9**, in pure $F_{x_{forceps}}$ loading situation, there are still obvious residual calculated values of $F_{y_{forceps}}$. And the waveform of $F_x$ is familiar with $F_y$, which proves these decoupled signals are not well uncorrelated. Thus, the errors of $F_{y_{forceps}}$ are amplified in calculation. The calculated values of $F_{z_{forceps}}$ will be with more errors.

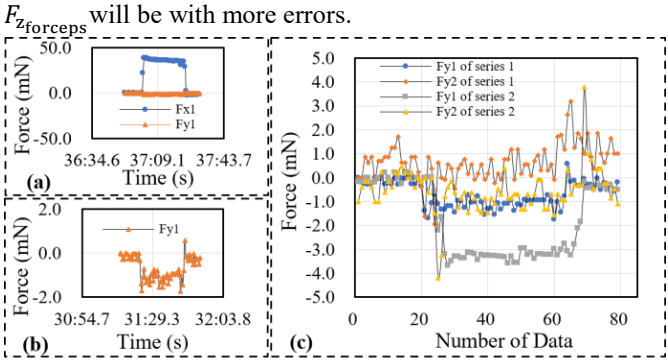

**Fig. 9.** The results of planar force with only exerted x-axial force of several samples: (a) The calculation of planar forces of one sample. (b) The waveform of $F_{y1}$. (c) The waveform of $F_{yi}$ of serval series.

### B. Components of signals

The raw signals from each sensor contain information corresponding to each direction of force information, noise from the environment, and system instability. Samples of system instability are shown in **Fig. 10**. The instability of the system is $\pm 2 \sim 5 \ pm$ during measurement as in **Fig. 10**. This kind of signal is much smaller than components corresponding to forces. The components of signals from sensors can be diagramed in **Fig. 10**. Noise from the environment comes from the influence of varying temperatures, vibrations, and collision with objects. Without drastic fluctuations in temperature, the noise can be small and temporary. And the temperature can be controlled to be stable. Based on the practice, the vibration from the operator's hands can hardly cause large deformations on prongs of forceps, and thus, the component from vibrations can be barely seen in the data. As for the collision between the sensors and soft objects, the deformation of sensors is limited, thus, the distribution of signals from sensors can be ignored.

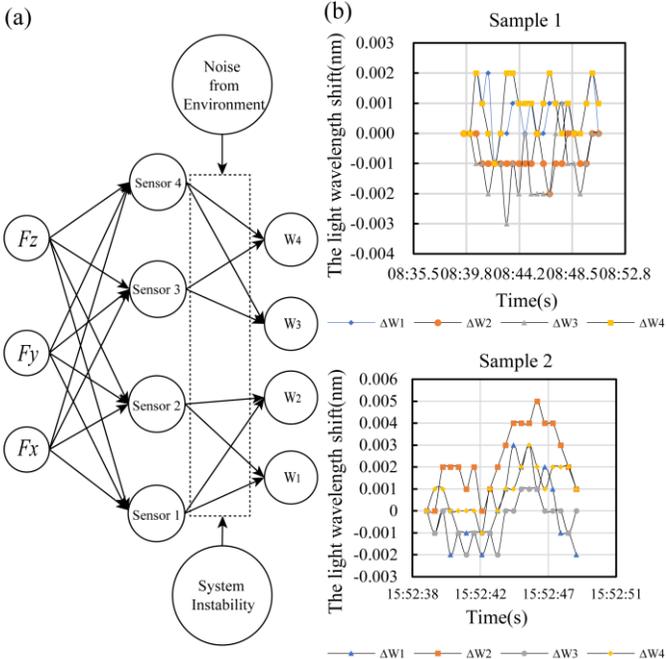

**Fig. 10.** (a) Illustration of the relationship between forces and components of signals. (b) Samples of signals corresponding to the system instability. The

The observed signals from sensors are composed of these elements in different proportions as (b) in **Fig. 10**. Without separating these elements from observed signals, the decoupled data will be mixed with components from these elements, because the mixed(coupling) matrix getting from the calibration matrix can decorrelate signals but is not precious enough, especially involving sensitive and wide-range measuring occasion. The shortage of the calibration method is listed below:

a. Limited by calibration equipment and human operation, the errors take more part in more previous and sensitive calibration occasions. The calibration process becomes inconvenient, time-consuming, and expensive.

b. The decoupling mainly depends on the combining matrix, the noise or errors contained in the matrix will bring calculation errors that cannot be ignored in signal analysis.

c. The noise cannot be separated and amplified as errors of forces in calculations.

### C. Method for improving signals processing performance combined with FastICA

There is a need to change the decoupling method to have a good estimate of the signals, because of the poor decoupling performance of using a coefficient matrix from the calibration process. It can be known that the inaccuracy comes from the matrix $\begin{bmatrix} k_{11} & k_{21} \\ k_{12} & k_{22} \end{bmatrix}$ and its inverse matrix. To avoid this, if the components of each sensor's signal correspond to each axial force, and the influence of $k_{21}$, $k_{12}$ can be avoided. The procedures of calibration and correction will be simplified because only $k_{11}$, $k_{12}$ adjusted rather than all four coefficients being adjusted.

Inspired by the idea of ICA: separating independent components from the observed signals by estimating characteristics of data, which is related to the independency of data, such as the negentropy. The FastICA is naturally applied in this paper to separate components of the observed signals from sensors. As mentioned in the previous sections, some preconditions should be satisfied for using FastICA:

1. The relationship between axial force and signals follows a linear relationship.

2. The axial forces on the XYZ axes are independent of each other.

3. The different sensor is sensitive to the different axial component of force.

4. The number of independent components can be equal to the number of sensors.

For the problem discussed in this paper, these conditions can be satisfied. The ICA is about separating linearly mixed independent components from the observed components, and FastICA is the fast form of ICA. For the observed signals $\begin{bmatrix} \Delta w_1 \\ \Delta w_2 \\ \Delta w_3 \\ \Delta w_4 \end{bmatrix}$, recovering the forces $\begin{bmatrix} F_x \\ F_y \\ F_z \end{bmatrix}$ means separating all the responses of sensors from the axial components of exerted forces. As in the equations in (12):

$$\begin{cases} \Delta w_1 = k_{11}F_{x1} + k_{12}F_{y1} + n_{ins} + n_{enviran} \\ \Delta w_2 = k_{21}F_{x1} + k_{22}F_{y1} + n_{ins} + n_{enviran} \\ \Delta w_3 = c_{11}F_{x2} + c_{12}F_{y2} + n_{ins} + n_{enviran} \\ \Delta w_4 = c_{21}F_{x2} + c_{22}F_{y2} + n_{ins} + n_{enviran} \end{cases} \quad (12)$$

Assuming $n_{enviran}$ is zero without significant fluctuations, each $\Delta w_i$ is the result of linear mixed wavelength shifts of sensor-i from the independent component of $\begin{bmatrix} F_{xi} \\ F_{yi} \\ F_z \end{bmatrix}$. Therefore, recovering the forces is the same as separating the light wavelength $\Delta w_{F_{xi}}$ and $\Delta w_{F_{yi}}$.

$$\begin{cases} \Delta w_1 = \Delta w_{F_{x1}} + \Delta w_{F_{y1}} + n_{ins} \\ \Delta w_2 = \Delta w_{F_{x1}} + \Delta w_{F_{y1}} + n_{ins} \\ \Delta w_3 = \Delta w_{F_{x2}} + \Delta w_{F_{y2}} + n_{ins} \\ \Delta w_4 = \Delta w_{F_{x2}} + \Delta w_{F_{y1}} + n_{ins} \end{cases} \quad (13)$$

In ICA, the independent sources $s$ become non-independent after linear coupling by A:

$$x = As \quad (14)$$

Where $x$ is the observed signals from sensors, A is the mixing matrix, and s is the independent variables. For getting the source s from x, the unmixing matrix should be ensured:

$$Y = W^T BAs \quad (15)$$

Where Y is the estimate of s, and B is the whitening matrix. The method to ensuring $W^T$ is to find the solution of minJ(y)= minJ($W^T Z$). The $W$ is the unmixing matrix for recovering the sources s. As illustrated above, the W is a $4 \times 4$ matrix as below:

$$W = \begin{bmatrix} a & b \\ c & d \end{bmatrix} \quad (16)$$

The W is the key to recovering the $\Delta w_{F_{ij}}$. However, because of the inherent ambiguity of FastICA, recovering the real amplitude of $\Delta w_{F_{ij}}$ just by FastICA cannot be realized. More restrictions are necessary to determine the true value of W. Some restrictions come from the reality of the scenarios:

1. All signals share the same units.
2. The energies of signals correspond to the magnitude of real forces.
3. The signs of signals from sensors reflect the directions of forces.

The real fluctuation of $\Delta w_{F_{ij}}$ can be well recovered by FastICA. Therefore, $\Delta w_{F_{ij}}$ can be recovered if the real value of unmixing matrix W can be recovered. Considering the real restrictions of the scenario, the variable $\Delta w_{F_{xi}}$ and $\Delta w_{F_{yi}}$ share the same units with the $\Delta w_{F_i}$, this means the real unmixing W can be described in (17):

$$W_{re} = \begin{bmatrix} 1 & \hat{b} \\ \hat{c} & 1 \end{bmatrix} \quad (17)$$

Therefore, the ambiguity of amplitude can be solved. In conclusion, the method for signal processing is to recover the real components of the unmixing matrix $W$ calculated by FastICA with some data preprocessing and reality conditions. The procedure of the method is listed below:

1. Partially filtering instability of system from signals.

**Algorithm 1** Filtering instability of the system partially
1: **Input: Times sequence of signals from sensors-$\Delta w_1$, $\Delta w_2$, $\Delta w_3$, $\Delta w_4$.**
2: **Output: Signals are partially filtered out $n_{ins}$ : $\Delta w_1'$, $\Delta w_2'$, $\Delta w_3'$, $\Delta w_4'$.**
3: **For** k = 1 to 4 do
4:   **For** i = 1 to max number of $\Delta w_k$-1 **do**
5:     j = i +1;
6:     **If** the absolute value of $\Delta w_k(i)$ is not larger than 0.002 nm **then**
7:       $\Delta w_k(i)' \leftarrow \Delta w_k(i)$ =0;
8:     **else**
9:       $\Delta w_k(i)' \leftarrow \Delta w_k(i)$;
10:     **End**
11:     **If** the absolute value of $\Delta w_k(j)$ is not larger than 0.002 nm **then**
12:       $\Delta w_k(j)' \leftarrow \Delta w_k(j)$ =0;
13:     **else**
14:       $w_k(j)' \leftarrow \Delta w_k(j)$;
15:     **End**
16:     **If** the absolute value of $\Delta w_k(j)$ & the absolute value of $\Delta w_k(j)$ is larger than0.002 pm **then**
17:       **If** the absolute value of the difference between the absolute value of $\Delta w_k(i)$ and the absolute value of $\Delta w_k(j)$ is larger than 0.002 pm **then**
18:         $\Delta w_k(i)' \leftarrow \Delta w_k(i)$, $\Delta w_k(j)' \leftarrow \Delta w_k(j)$;
19:       **else**
20:         $\Delta w_k(i)' \leftarrow \Delta w_k(i)$, $\Delta w_k(j)' \leftarrow \Delta w_k(i)$;
21:       **End**
22:     **End**
23:   **End**
24: **End**

2. Ensuring the direction of axial force.

Table III
THE RELATIONSHIP BETWEEN SIGNS OF SIGNALS AND DIRECTIONS OF FORCES

| Sensor | Signs of signals | Axial force |
|---|---|---|
| 1 | + | Negative $F_{x1}$ |
| | - | Positive $F_{x1}$ |
| 2 | + | Negative $F_{x2}$ |
| | - | Positive $F_{x2}$ |
| 3 | + | Negative $F_{y1}$ |
| | - | Positive $F_{y1}$ |
| 4 | + | Negative $F_{y2}$ |
| | - | Positive $F_{y2}$ |

3. Ensuring axial components of force:

It is necessary to distinguish the uniaxial force exerted and the multiple axial forces exerted situation.

**Algorithm 2 Ensuring axial forces are mixed or not**
**Input:** $\Delta w_1'$, $\Delta w_2'$, $\Delta w_3'$, $\Delta w_4'$.
**Output:** uniaxial force exerted or multiple axial forces exerted.
1: **For** k = 1 to 3 **do**
2:   Using FFT to decompose the $\Delta w_k'$, $\Delta w_{k+1}'$ and $\Delta w_{k\rightarrow fft}'$, $\Delta w_{k+1\rightarrow fft}'$can be obtained
3:   Removing the largest 2-3 values of $\Delta w_{k\rightarrow fft}'$, $\Delta w_{k+1\rightarrow fft}'$ and new values $\Delta w_{newk\rightarrow fft}'$, $\Delta w_{newk+1\rightarrow fft}'$are obtained:
4:   doing the linear regression of $\Delta w_{newk\rightarrow fft}'$ and $\Delta w_{newk+1\rightarrow fft}'$ and get the $R^2$;
5:   **If** $R^2 > 0.8$ **then**
6:     uniaxial force exerted

| 7: | **Else** |
|---|---|
| 8: | Multiple axial forces exerted |
| 9: | **End** |
| 10: | k = k+2 |
| 11: | **End** |

4. Separating signals by FastICA and recovering the components of signals with the restrictions of the scenario (using recovered unmixing matrix W):

**Algorithm 3 Recovering components of signals**

**Input:** $\Delta w_1'$, $\Delta w_2'$, $\Delta w_3'$, $\Delta w_4'$.

**Output:** components of signals from each sensor $\Delta w_{fxi}$, $\Delta w_{fyi}$, $n_{ins}$.

| 1: | **For** k = 1 to 3 **do** |
|---|---|
| 2: | Getting new array $m \leftarrow (\Delta w_k', \Delta w_{k+1}')$; |
| 3: | Whiting the centralization of m and getting the $m_{whiting}$; |
| 4: | Ensuring the number of independent variables is 2; |
| 5: | **For** i = 1 to max number of independent variables **do** |
| 6: | Initializing the vector-matrix w by a random number, $a_2 \leftarrow 1$, $u \leftarrow 1$; |
| 7: | Normalizing the w; |
| 8: | Optimizing the function: $J_G(w) = [E\{G(w^T x)\} - E\{G(v)\}]^2$, the restriction is $\|w\|^2 = 1$, $G(u) = -\frac{1}{a_2} a_2^{-\frac{u^2}{2}}$; |
| 9: | **Do iterations** |
| 10: | $w_{n+1} = E\{xG'(w_n^T x)\} - E\{G''(w_n^T x)\}]w_n$ |
| 11: | $w_{n+1} = \frac{w_n}{\|w\|^2}$ |

| 12: | **Until** the maximum number of iterations is reached; |
|---|---|
| 13: | $w$ is saved as one component of W; |
| 14: | **End** |
| 15: | Ensuring the arrangement of w according to the arrangements of signals, the w should recover the data in the sequence corresponding to the $\Delta w_1'$, $\Delta w_2'$, $\Delta w_3'$, $\Delta w_4'$ |
| 16: | **Return** $W \leftarrow w$; |
| 17: | **Return** $W_{re}$ |
| 18: | Recovering components of signals by $W_{re}$ |
| 19: | k = k+2 |
| 20: | **End** |
| 21: | **For** uniaxial force exerted do |
| 22: | **Return** $\Delta w_{fxi}$ or $\Delta w_{fyi}$, $n_{ins}$; |
| 23: | **End** |
| 24: | **For** multiple axial forces exerted |
| 25: | **Return** $\Delta w_{fxi}$, $\Delta w_{fyi}$; |
| 26: | **End** |

5. Calculating the value of forces.

The forces can be calculated according to $\Delta w_{fxi}$, $\Delta w_{fyi}$ by (26), commonly the $F_{x1} \approx F_{x2}$ in most situations:

$$\begin{bmatrix} F_{x1} \\ F_{y1} \\ F_{x2} \\ F_{y2} \\ n_{ins} \end{bmatrix} = \begin{bmatrix} a_{11} & 0 & 0 & 0 & 0 \\ 0 & a_{22} & 0 & 0 & 0 \\ 0 & 0 & a_{33} & 0 & 0 \\ 0 & 0 & 0 & a_{44} & 0 \\ 0 & 0 & 0 & 0 & 1 \end{bmatrix} \begin{bmatrix} \Delta w_{fx1} \\ \Delta w_{fy1} \\ \Delta w_{fx2} \\ \Delta w_{fy2} \\ n_{ins} \end{bmatrix} \tag{18}$$

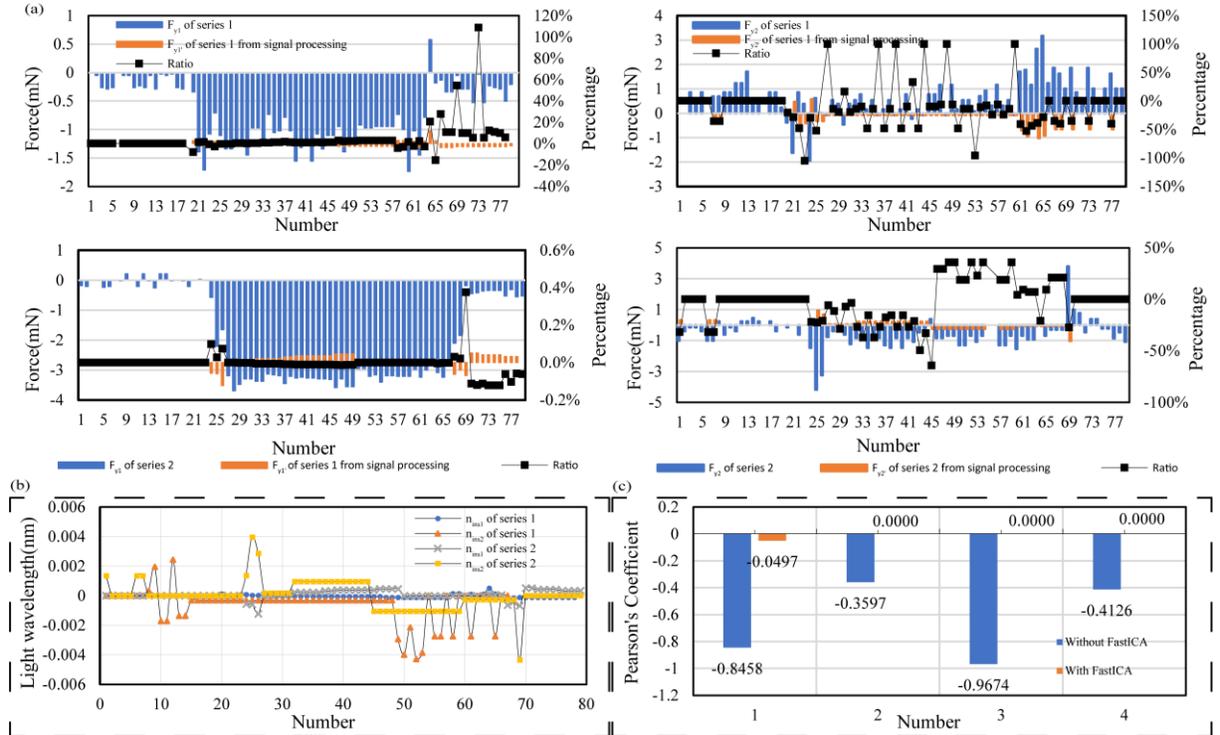

**Fig. 11.** (a) Contrast between results calculated by the method using FASTICA and the matrix obtained from calibration. (b) The recovered $n_{ins}$ of samples. (c) The Pearson's coefficient of recovered results from the method with FastICA and without FastICA.

### D. Results of signal processing combined with FastICA

The method based on FastICA has a better performance in separating components of signals than directly using the decoupling matrix obtained from the calibration. The advantages of this method are simplifying the calibration and correcting process:

1. It is the $k_{11}$, $k_{22}$ in the $\begin{bmatrix} k_{11} & k_{21} \\ k_{12} & k_{22} \end{bmatrix}$ and $c_{11}$, $c_{22}$ in the $\begin{bmatrix} c_{11} & c_{21} \\ c_{12} & c_{22} \end{bmatrix}$ that have to be calibrated in the method based on the FastICA. This makes the calibration and compensation process more simplified to operate.

2. Components can be well separated:

The ICA is a strong method to recover independent components, so the correlation among signals can be well weakened. For the only x-axial force exerted, the components of $\Delta w_1$ and $\Delta w_2$ can be described in (19):

$$\begin{bmatrix} \Delta w_1 \\ \Delta w_2 \end{bmatrix} = \begin{bmatrix} k_{11} & 1 \\ k_{12} & 1 \end{bmatrix} \begin{bmatrix} F_{x1} \\ n_{ins} \end{bmatrix} \tag{19}$$

The $n_{ins}$ of samples (**Fig. 9**) obtained after signal processing ranges from $\pm 5 \, pm$, which is corresponding to the facts and indicates that the signals are well recovered(**Fig. 11** (b)).

Results of the samples of uniaxial force exerted recovered by the method combined with FastICA are shown in **Fig. 11**, the values of $F_{y_{forceps}}$ calculated by the method combined with FASTICA are more correspond to the true values(with errors decreasing up to 50% least )and the shape of the curve is not familiar with the waveform of $F_{x_{forceps}}$(less correlated). Therefore, components can be well separated.

## V. EXPERIMENT SETUP AND RESULTS

### A. Calibration setup and result

Each sensor installed on the forceps is sensitive to different axial components of forces and all sensors can have a more or less response from exerted forces according to mechanics and simulations. For estimating the relationship between signals of sensors and the exerted forces. The sensors on forceps are calibrated on a calibration platform as shown in **Fig. 12**.

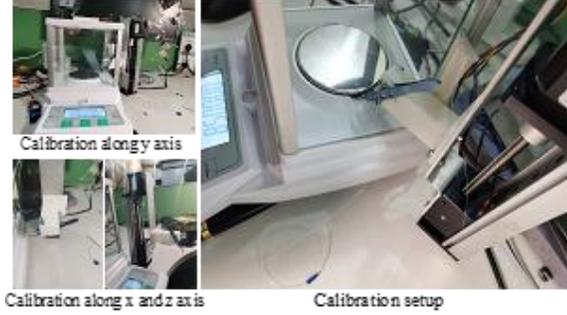

**Fig. 12.** Setup for calibration.

As shown in **Fig. 5**, the fixed areas have a tiny influence on the occurred strains. The calibration platform, including the step displacement platform, digital electronic scales, and a manipulator with a start-stop switch was constructed to calibrate the sensor. The calibration results are shown in **Fig. 7** and **Fig. 13**.

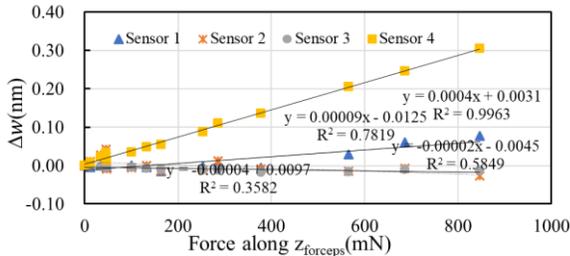

**Fig. 13.** Calibration results for sensors with $F_{z_{forcpes}}$ exerted. The $\Delta w$ is the light wavelength shifts.

From **Fig. 7**, the relationship between the light wavelength shifts and forces is linear. Different sensors are sensitive to the particular axial components of forces. $F_{z_{forceps}}$ is special for Sensor-1,2, and 3 as shown in **Fig. 13**. The responses from Sensor-1,2, and 3 correspond to $F_{z_{forceps}}$ is small, irregular and not linear. For $F_{z_{forceps}}$, only Senor 4 is sensitive to $F_{z_{forceps}}$, the response from Sensor 4 is significantly larger than other sensors, which corresponds to the fact that only Sensor 4 can detect the $F_{z_{forceps}}$ because the compression of metal are not large enough to be easily observed by sensors. The calibration equations can be obtained from **Fig. 7**:

$$\begin{bmatrix} \Delta w_1 \\ \Delta w_2 \end{bmatrix} = \begin{bmatrix} 0.0036 & 0.00009 \\ 0.0004 & -0.0042 \end{bmatrix} \begin{bmatrix} F_{x1} \\ F_{y1} \end{bmatrix} + \begin{bmatrix} 0.0071 \\ -0.0032 \end{bmatrix} \tag{15}$$

$$\begin{bmatrix} \Delta w_3 \\ \Delta w_4 \end{bmatrix} = \begin{bmatrix} 0.0016 & 0.0003 \\ 0.0006 & -0.0015 \end{bmatrix} \begin{bmatrix} F_{x2} \\ F_{y2} \end{bmatrix} + \begin{bmatrix} 0.0025 \\ -0.0037 \end{bmatrix} \tag{16}$$

Where the $\Delta w_i$ stands for the light wavelength shifts of sensor-$i$. and forces can be calculated by inversing the above equations:

$$\begin{bmatrix} F_{x1} \\ F_{y1} \end{bmatrix} = \begin{bmatrix} 277.1 & 5.9 \\ 26.4 & -237.5 \end{bmatrix} \begin{bmatrix} \Delta w_1 \\ \Delta w_2 \end{bmatrix} - \begin{bmatrix} 1.95 \\ 0.95 \end{bmatrix} \tag{17}$$

$$\begin{bmatrix} F_{x2} \\ F_{y2} \end{bmatrix} = \begin{bmatrix} 581.4 & 116.3 \\ 232.6 & -620.2 \end{bmatrix} \begin{bmatrix} \Delta w_3 \\ \Delta w_4 \end{bmatrix} - \begin{bmatrix} 1.02 \\ 2.88 \end{bmatrix} \tag{18}$$

The $F_{z_{forceps}}$ can be estimated as below:

$$F_{z_{forceps}} = \frac{F_{y2} - k_n F_{y1} \cos\alpha}{\sin\alpha} \approx 4.1 (F_{y2} - F_{y1}) \tag{19}$$

### B. Set up for Experiments on thin sheets and result

The setup for the experiment is fixing a sample of the thin sheet (material: PO ) on a smooth plane (**Fig. 14**).

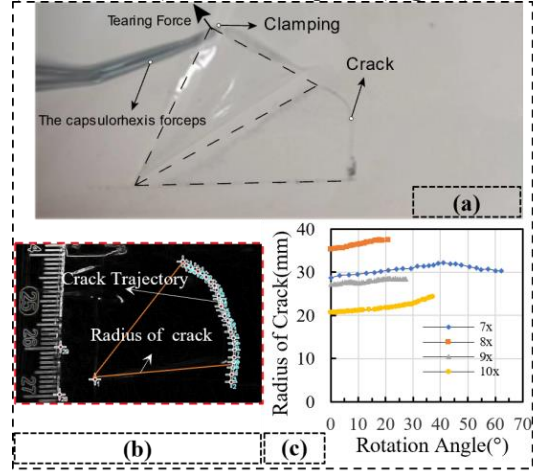

**Fig. 14. (a)** Experiment setup for tearing the thin sheets. (b) Crack propagation on PO thin sheet tearing by using forceps. (c) Samples of the radius of the crack in tearing experiments.

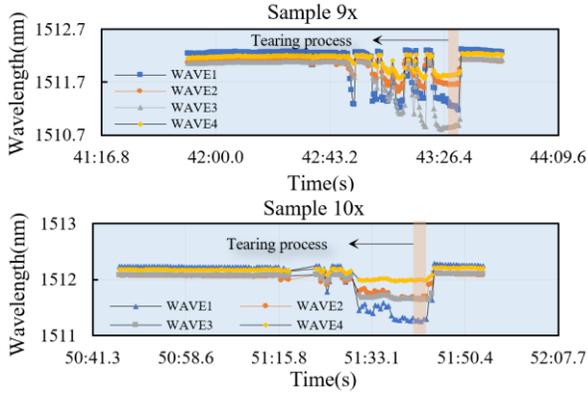

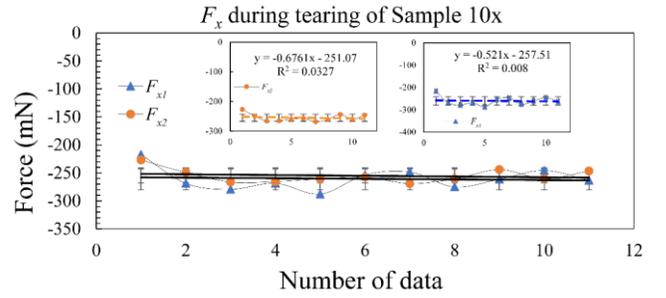

**Fig. 17.** Samples of $F_{x1}$ and $F_{x2}$ in the tearing process

**Fig. 15.** Samples of signals from sensors during the tearing of the thin sheet.

From **Fig. 15**, the interaction forces can be recorded by sensors. It can be seen signals from the sensor are more stable in the tearing process than signals from other periods in some operations. This phenomenon may indicate that the tearing forces can be stable during the propagation of circular cracks.

### C. Recovering data from the experiment

For the experiments on the PO sheets, the results of forces are shown in **Fig. 16** and **Fig. 17**.

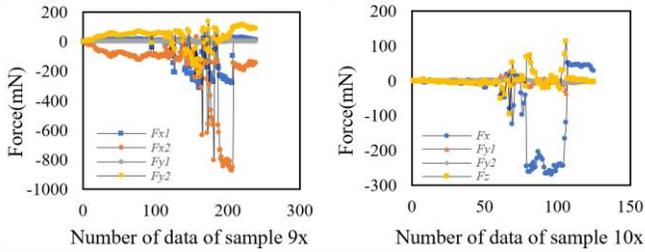

**Fig. 16.** Samples of forces in the experiments on thin sheets.

For sample 9x, the $F_{x1}$ is not close to $F_{x2}$, this is because the bulge does not work to counteract the excessive pressures from hands. Therefore, $F_{x2}$ of sample 1 is larger than $F_{x1}$ of sample 1 in **Fig. 17**.

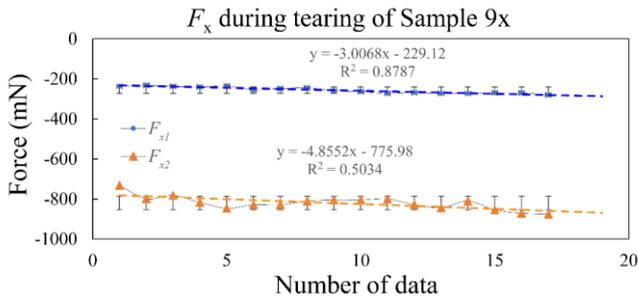

TABLE IV
FORCES ALONG X-AXIS MEASURED BY SENSORS DURING TEARING ON THIN SHEETS

| Force | Mean ± SD (mN) | COV | $cot^{-1}\frac{F_{xi}}{F_{yi}}$ (°) | P value of L-B test | P value of KPSS test | $P_i$ of ADF test |
|---|---|---|---|---|---|---|
| $F_{x1}$ of sample 9x | 256.18±16.2 | 0.063 | 0.56±0.44 | 0.002 | 0.094 | $P_1$=0.03 |
| $F_{x2}$ of sample 9x | 819.68±34.56 | 0.042 | 2.77±1.09 | 0.02 | 0.074 | $P_1$=0.00 |
| $F_{x1}$ of sample 10x | 260.63±19.35 | 0.074 | 0.55±0.37 | 0.0001 | 0.076 | $P_1$=0.00 |
| $F_{x2}$ of sample 10x | 255.12±12.4 | 0.048 | 0.16±0.16 | 0.00005 | 0.088 | $P_0$=0.005 |

As in **Fig. 17,** the coefficients of variation of samples are generally below 0.1, which indicates that $F_{x2}$ and $F_{x1}$ are stable during the propagation of the crack. As in Table IV, the angles between the vector $F_{xi}$ and $F_{yi}$ are 0.56±0.44°, 2.77±1.09°, 0.55±0.37°, and 0.16±0.16°, separately. Thus, it proves that the $F_{xfroceps}$ play the main role in the resultant force during crack propagation. And the $F_{xfroceps}$ is always perpendicular to the radius of the crack as shown in **Fig. 14**. The $P_i$ is the factor obtained from the Augmented Dickey-Fuller Test (ADF test). From the small enough $P_i$ in Table IV, it can be seen that the time sequences of $F_{x1}$ and $F_{x2}$ in sample 9x, $F_{x1}$ in sample 10x are corresponding to first-order difference stationary process. The time sequence of $F_{x1}$ in sample 10x shows its stationary process. As for $F_{x1}$ and $F_{x2}$ in sample 9x and $F_{x1}$ in sample 10x, they are not stationary for the existence of trend term. After excluding the trend term, these sequences become stationary. Generally, the forces during the tearing process show a very interesting property, they are most well corresponding to the stationary process. It indicates that the forces in circular propagation are likely to be stable. Combined with the crack radius data in **Fig. 14** (c), it indicates that stable forces can be one of the factors that determine a circular crack or one of the outcomes of a circular crack in some situations. However, the exact relationship between the forces and the circular propagating crack is not known yet.

## VI. CONCLUSION

In this paper, the mechanics of capsulorhexis forceps with forces exerted on tips have been modeled and discussed. Based on the mechanics model, a three-dimensional force perception method on capsulorhexis forceps was first developed by installing FBGs in the designed position and sequence. The resolution on the x, y, and z-axis of this sensing method is 0.5 mN, 0.5 mN, and 2 mN separately, which provide a sensitive enough method to measure the forces exerted on tips of forceps in CCC. The incomplete decoupling phenomenon in the commonly used calibration method is first found and discussed in this paper. This phenomenon can become obvious for sensitive measuring occasions because the results of each axial force are unavoidably mixed with the influence of the other two axes without a strong decoupling method. It is found that this coupling effect can still leave considerable residuals even after calibration for sensitive measurement occasions. For better and more accuracy, a method combined with FastICA is proposed, discussed, and firstly applied in this paper for processing the signals from sensors and recovering them to be closer to real forces. This method can significantly reduce the residuals of forces that happened in calculation by weakening the coupling among signals. This method can be adjusted to all the multi-sensor measuring calibration and this method can simplify the procedures for calibration and compensation of the signals. Furthermore, an interesting phenomenon for circular crack propagation during tearing is found. Based on the results of our experiments, the time sequence of tearing forces seems to be stable and regular without fierce fluctuations rather than irregular changes, which cannot be tracked. This phenomenon indicates some physical variables are responsible for circular crack propagation during tearing. Further research is needed to explore the kind of relationship behind this phenomenon.